\documentclass[11pt]{article}

\usepackage[preprint]{acl}

\usepackage{times}
\usepackage{latexsym}

\usepackage[T1]{fontenc}
\usepackage[utf8]{inputenc}

\usepackage{microtype}

\usepackage{inconsolata}

\usepackage{graphicx}

\usepackage{booktabs}
\usepackage{multirow}
\usepackage{makecell}
\usepackage{xspace}
\usepackage{graphicx}
\usepackage{enumitem}
\usepackage[most]{tcolorbox}
\usepackage{tabularx}
\usepackage{enumitem}
\usepackage[table]{xcolor}

\usepackage[most]{tcolorbox}
\usepackage{fancyvrb}

\tcbuselibrary{breakable}
\usepackage{listings}

\newtcolorbox[list inside=prompt]{prompt}[1][]{
    colbacktitle=black!60,
    coltitle=white,
    fontupper=\footnotesize,
    boxsep=5pt,
    left=0pt,
    right=-1pt,
    top=0pt,
    bottom=0pt,
    boxrule=1pt,
    width=\textwidth,
    breakable,
    #1,
}

\title{Towards Verifiable Multimodal Deep Research: A Multi-Agent Harness for Interleaved Report Generation}

\author{Chenghao Zhang, Guanting Dong, Yufan Liu, Tong Zhao, Xiaoxi Li, Zhicheng Dou\thanks{Corresponding author.} \\
Gaoling School of Artificial Intelligence, Renmin University of China \\
\texttt{davidzhang@ruc.edu.cn, dou@ruc.edu.cn} \\
}

\begin{document}
\maketitle
\begin{abstract}
Large Language Models (LLMs) have advanced autonomous agents from deep search, which retrieves concise factual answers, to deep research, which synthesizes scattered evidence into long-form reports.
However, verifiable multimodal deep research remains challenging due to open-ended synthesis without deterministic ground truth and the need to interleave textual arguments with visual evidence.
We propose \textsc{Ptah}, a multi-agent harness for interleaved report generation.
\textsc{Ptah} orchestrates the lifecycle from user query to rendered web report through planning, research, and writing stages, where specialized agents construct visual-aware plans, collect claim-grounded evidence, maintain source-aligned images in a Visual Working Memory, and compose reports through declarative multimodal tool use.
A verifier agent serves as the harness's acceptance function, enforcing factual grounding, citation fidelity, and cross-modal consistency throughout the workflow.
We further introduce \textsc{Ptah}Eval, an evaluation protocol that augments existing benchmarks with image-level and presentation-level assessments.
Experiments on deep research benchmarks show that \textsc{Ptah} produces more reliable, visually informative, and usable human-facing multimodal reports than strong baselines. Our code is released at \url{https://github.com/SnowNation101/Ptah}
\end{abstract}

%% Introduction
\section{Introduction}

%% Background: LLM/VLM to RAG to Deep Search to Deep Research
In recent years, Large Language Models (LLMs)~\cite{tongyi2025qwen3, qwen2025qwq32b, ds2025dsv32} and Vision-Language Models (VLMs)~\cite{bai2025qwen3vl, qwen2026qwen35} have demonstrated exceptional reasoning capabilities in content understanding and generation, enabling them to tackle sophisticated, cross-domain challenges. 
However, the inherent issue of hallucination remains a critical bottleneck for their deployment in knowledge-intensive tasks. 
To mitigate this, Retrieval-Augmented Generation (RAG)~\cite{gao2023ragsurvey, zhang2025nyx, dong2025dparag} has emerged as a prevailing paradigm, leveraging external knowledge bases and search tools to provide factual grounding. 

\begin{figure}[t]
    \centering
    \includegraphics[width=\linewidth]{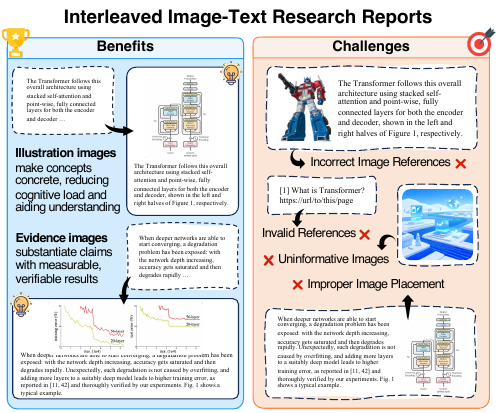}
    \vspace{-0.5em}
    \caption{Illustration of how images enhance report quality, and the challenges of generating high-quality interleaved image–text reports.}
    \label{fig:intro}
\end{figure}

Building on this paradigm, \textit{Deep Search} has emerged across both academia and industry as an agentic multi-step search paradigm, where autonomous agents leverage complex toolchains to tackle more demanding tasks.
Benchmarks such as GAIA~\cite{mialon2024gaia} and HLE~\cite{phan2025hle}, along with complex mathematical reasoning tasks, have showcased the efficacy of multi-step search and reasoning in solving hard problems. 
Nevertheless, these tasks are primarily characterized by \textit{deterministic answers} in closed domains, where outcomes can be rigorously verified and refined through ground-truth labels or automated scripts.

%% Large scale problems
In contrast, the recent emergence of \textit{Deep Research} systems in industry (e.g., OpenAI Deep Research~\cite{openai2025}) marks a paradigm shift from seeking singular, objective answers toward synthesizing comprehensive, long-form reports. 
Compared with closed-end deep search, deep research poses two distinctive challenges:
\textbf{(1) Open-endedness.} Deep research reports lack a deterministic ground truth, requiring agents to perform multi-round iterative searches in open domains where outputs cannot be straightforwardly verified.
\textbf{(2) Multimodal interleaving.} A professional report characteristically interleaves text with visual evidence such as trend charts and illustrative figures (Figure~\ref{fig:intro}), demanding tight integration of multimodal content rather than text-only synthesis.

%% Specific problems
Despite the rapid progress of these systems, existing approaches fall short on both fronts. 
For \textit{open-endedness}, multi-step research pipelines lack stage-wise verification, allowing noise introduced early on to accumulate and ultimately produce factually unreliable text and misaligned visuals. 
For \textit{multimodal interleaving}, current frameworks treat image integration as a post-hoc decorative step rather than a core component of the research process, leaving visual evidence loosely tied to textual arguments and far from the interleaved quality expected in professional reports. 
These shortcomings motivate a holistic agentic approach that can autonomously plan, investigate, and verify research findings within a unified multimodal loop.

%% Our method
To address these challenges, we propose \textbf{\textsc{Ptah}}\footnote{Named after Ptah, the ancient Egyptian creator deity and patron of craftsmen, the name reflects the harness's role in orchestrating the composition of structured multimodal reports from heterogeneous textual and visual materials.}, an agentic harness for credible multimodal deep research.
Rather than treating multimodal report generation as a monolithic generation problem, \textsc{Ptah} organizes specialized agents, external tools, intermediate research states, and verification signals into a controlled execution workflow.
The harness orchestrates the full lifecycle from user query to rendered multimodal report through three stages: \textit{Planning}, \textit{Research}, and \textit{Writing}.
In \textit{Planning}, \textsc{Ptah} constructs a visual-aware research plan that specifies both textual structure and intended visual evidence.
In \textit{Research}, parallel agents instantiate this plan with claim-grounded evidence, citations, numerical data, and source-aligned visual candidates maintained as intermediate research state.
In \textit{Writing}, a writer agent composes the final interleaved report through declarative multimodal tool use.
Across all stages, verifier hooks serve as the harness's acceptance function, checking protocol compliance, factual grounding, citation fidelity, visual relevance, and cross-modal consistency before the workflow advances.

Furthermore, to bridge the gap in evaluation metrics for interleaved image--text reports, we introduce \textbf{\textsc{Ptah}Eval}, a flexible evaluation protocol that integrates seamlessly into existing deep research benchmarks. \textsc{Ptah}Eval assesses report quality along two dimensions: \textit{Image Content Quality} and \textit{Multimodal Presentation Quality}. Experimental results demonstrate that \textsc{Ptah} generates high-quality, credible, and professionally interleaved research reports.

To summarize, we make the following contributions:
\begin{itemize}[leftmargin=1em]
    \item We propose \textsc{Ptah}, an agentic harness that coordinates specialized agents, external tools, research states, and verification signals for credible multimodal deep research.

    \item We design a visual-aware workflow that organizes multimodal deep research into \textit{Planning}, \textit{Research}, and \textit{Writing}, maintaining plans, evidence, citations, numerical data, and source-aligned visual candidates as inspectable intermediate artifacts.

    \item We introduce verifier hooks that implement the harness's acceptance function, enabling stage-wise checks for protocol compliance, factual grounding, citation fidelity, visual relevance, and cross-modal consistency.

    \item We present \textsc{Ptah}Eval, an evaluation protocol for interleaved image--text research reports, and show that \textsc{Ptah} improves multimodal report quality and readability while maintaining strong textual reliability.
\end{itemize}
%%% End of Introduction

\section{Related Work}

\subsection{Deep Search and Deep Research}
Following ReAct~\citep{yao2023react}, deep search augments LLMs with iterative tool use for multi-step information retrieval. Early efforts extend RAG with iterative retrieval and evidence verification~\citep{press2023selfask,shao2023iter-retgen,asai2024selfrag}, while more recent work generalizes this into agent-based frameworks with richer action spaces~\citep{wang2024codeact,li2025searcho1,jin2025searchr1,chen2025research,wu2025webdancer,dong2025tool,dong2025arpo,dong2025aepo,dong2026agent}. However, these approaches primarily target closed-end question answering with deterministic answers~\citep{xi2025searchsurvey,wu2025webwalker}.

Recent systems extend deep search to open-ended, long-form report generation, including OpenAI Deep Research~\citep{openai2025}, Grok Deep Research~\citep{grok2025}, WebThinker~\citep{li2025webthinker}, OWL~\citep{hu2025owl}, Auto Deep Research~\citep{tang2025autodeepresearch}, and Multimodal DeepResearcher~\citep{yang2025multimodaldeepresearcher}. Nevertheless, most systems struggle to jointly achieve deep multi-hop reasoning and broad information coverage, exposing fundamental limitations of single-agent architectures in complex research settings~\citep{lan2025deepwidesearch,yen2025lost,shi2025deep}.

\subsection{Interleaved Image--Text Generation}
While recent MLLMs such as Qwen3-VL~\citep{bai2025qwen3vl}, InternVL~\citep{chen2023internvl}, GPT-4V~\citep{openai2023gpt4}, and LLaVA~\citep{liu2023llava} excel at understanding interleaved image–text inputs, they are primarily designed for perception and generally cannot generate interleaved outputs~\citep{deng2025bagel,xie2025show-o}.

Two paradigms have emerged for interleaved generation~\citep{guo2025llm-i}. The first builds native multimodal generative models within unified architectures, integrating diffusion-based decoders with autoregressive language models~\citep{xie2025show-o2,wu2025janus,meta2025chameleon,wu2024nextgpt,ge2024seedx,caffagni2024mllmsurvey}. The second treats interleaved generation as a tool-augmented agentic process, exemplified by THYME~\citep{zhang2025thyme} and WebWatcher~\citep{geng2025webwatcher}. Dedicated benchmarks such as MM-Interleaved~\citep{tian2024mminterleaved}, OpenLEAF~\citep{an2024openleaf}, and ISG-Bench~\citep{chen2025isgbench} further support evaluation of interleaved generation quality. However, existing methods generally lack explicit verification and cross-modal consistency checks, often producing weakly grounded visual outputs in open-ended scenarios.

\begin{figure*}[t]
    \centering
    \includegraphics[width=0.95\linewidth]{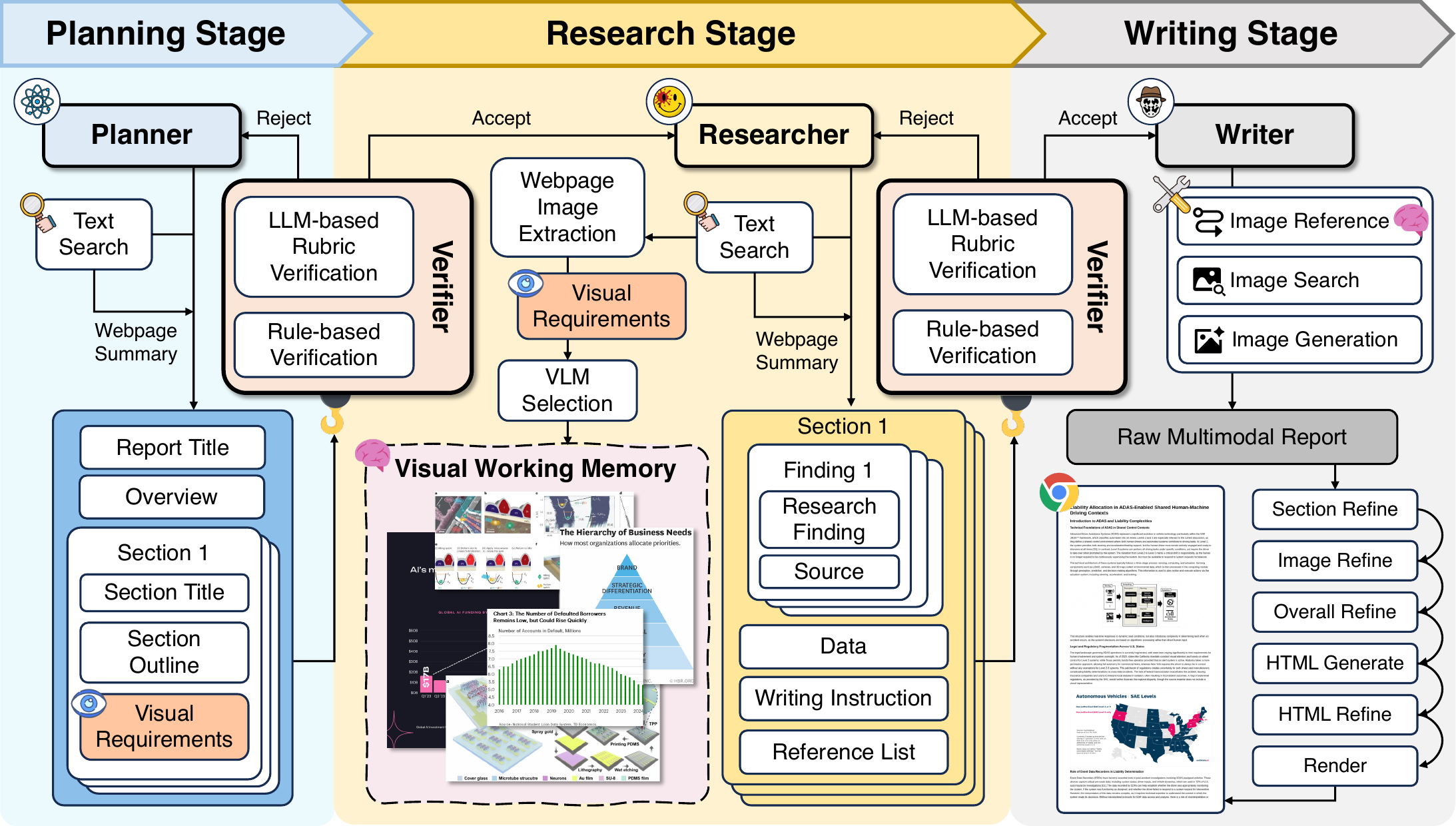}
    \vspace{-0.5em}
    \caption{Overview of \textsc{Ptah}, a multi-agent harness for verifiable multimodal deep research.}
    \vspace{-0.5em}
    \label{fig:main}
\end{figure*}

\section{Task Formulation}

Given a plain-text user query $q$, our goal is to produce a multimodal research report $r$ and its rendered web page $h$. We represent $r$ as an ordered sequence of content blocks
\begin{equation}
r = (b_1, b_2, \dots, b_M),
\end{equation}
where each block $b_i$ is either a textual segment $t_i$ or a visual element $v_i$, allowing flexible interleaved layouts such as $(t_1,v_1,v_2,t_2,\dots)$ that reflect the structure of research reports.

We formulate multimodal deep research as a harnessed agentic process. At step $t$, the harness maintains a research state $s_t = (q, \mathcal{M}_t, \tau_{<t})$, where $\mathcal{M}_t$ is the structured working state---intermediate plans, evidence, citations, numerical data, and visual candidates---and $\tau_{<t}$ is the interaction history. The model produces a reasoning step $z_t$ and may invoke a tool $a_t = (u_t, x_t)$ with $u_t \in \mathcal{U}$, yielding an observation $o_t = u_t(x_t)$ that updates $\mathcal{M}_{t+1}$; we write $\tau = \{(z_t, a_t, o_t)\}_{t=1}^{T}$ for the full trajectory.

After the research state is constructed, the final report is sampled as $r \sim p_{\theta}(\cdot \mid q, \mathcal{M}_T, \tau)$ and rendered into the final web page $h = \operatorname{Render}(r)$, where $\operatorname{Render}(\cdot)$ first serializes the interleaved blocks into HTML and then displays them as a webpage.

%% Ptah Method
\section{\textsc{Ptah}: Verifiable Multi-Agent Harness}

\textsc{Ptah} is an agentic harness for credible multimodal deep research.
As illustrated in Figure~\ref{fig:main}, it orchestrates the lifecycle from a user query to a rendered multimodal report through three stages: \textit{Planning}, \textit{Research}, and \textit{Writing}.
The \textit{Planner Agent} constructs a visual-aware research plan, the \textit{Researcher Agents} instantiate it with claim-grounded evidence and source-aligned images stored in \textit{Visual Working Memory}, and the \textit{Writer Agent} composes the final interleaved report through declarative multimodal tool use.
Across this lifecycle, a \textit{Verifier Agent} acts as the harness's acceptance function, combining rule-based checks with LLM-based rubric verification to ensure protocol compliance, factual grounding, citation fidelity, visual relevance, and cross-modal consistency before the workflow advances.

\subsection{Planning: Visual-Aware Research State Initialization}

The \textit{Planner Agent} initializes the research state by iteratively invoking text search tools to explore relevant domain knowledge.
It produces a structured plan that contains a high-level overview, section-level research goals, expected evidence types, and explicit visual specifications.
These visual specifications describe where visual elements should appear, what communicative role they should serve, and which form of visual evidence, such as charts, diagrams, screenshots, or illustrative figures, best supports the narrative.

The plan acts as the first structured working state maintained by the harness.
It constrains downstream research and writing by making the expected textual coverage and visual evidence explicit. Once produced, the plan is checked by the \textit{Verifier Agent} on two levels: rule-based validation of the interaction protocol, tool-use constraints, and JSON format; and LLM-based rubric assessment of query coverage, section coherence, and visual--argument relevance. Plans that fail either check are revised, optionally with additional searches, before the workflow proceeds.

\subsection{Research: Parallel Evidence Collection and Visual Working Memory}

While the planning stage determines the breadth of the report, the research stage instantiates the plan with grounded evidence.
For each planned section, a \textit{Researcher Agent} performs an independent investigation through search and retrieval tools.
Each researcher produces a structured research package containing key findings, claim-grounded evidence, numerical data, tables, references, and writing instructions for the downstream writer.
This design allows the harness to scale the research process across sections while keeping each section's evidence traceable and inspectable.

In parallel with textual evidence collection, each researcher extracts images from visited webpages and constructs a task-specific \textit{Visual Working Memory}.
Here, visual evidence is broadly defined as source-aligned visual material that supports, explains, or contextualizes the report content, including charts, screenshots, diagrams, photographs, and illustrative figures.
Raw image candidates first undergo rule-based filtering to remove low-resolution, duplicate, irrelevant, or non-informative images.
Then, a VLM-based selector evaluates the remaining candidates according to the visual requirements specified in the planning stage.
Each retained visual candidate is stored together with its source URL, surrounding webpage context, associated section, and intended visual role.
By externalizing webpage images into \textit{Visual Working Memory}, \textsc{Ptah} preserves source-aligned visual materials as structured cross-modal state rather than treating images as post-hoc decorative assets.

Each research package is then checked by the \textit{Verifier Agent} for citation, including claim support, coverage of the planned goals, numerical/reference consistency, and visual relevance to the section intent. Packages that fail are returned to the corresponding researcher for revision or further evidence collection.
\subsection{Writing: Declarative Multimodal Composition}

The \textit{Writer Agent} composes the report using the global plan, verified research packages, and \textit{Visual Working Memory}.
Instead of selecting and inserting images through an ad hoc post-processing step, the writer follows a declarative multimodal composition strategy.
It generates textual content and image directives jointly, embedding image tool tags at the positions where visual elements should appear.
These specify the intended visual role and the tool operation required to realize the image.

The harness then arbitrates among three types of image operations.
\textit{Image Reference} reuses source-aligned images from \textit{Visual Working Memory} and is preferred when suitable candidates exist.
\textit{Image Search} retrieves additional web images when the existing \textit{Visual Working Memory} does not satisfy the section requirement.
\textit{Image Generation} creates new visual elements when the report requires synthesized visuals, such as charts, structured diagrams, or illustrative figures.
For data-driven visuals, \textsc{Ptah} can invoke executable code rendering to generate charts or visualizations; for illustrative content, it can invoke image generation models from textual descriptions.
After all sections are composed, the writer generates a conclusion and assembles the sections into a raw interleaved report.

\paragraph{Test-Time Scaling}
After initial composition, \textsc{Ptah} applies verifier-guided test-time scaling through a sequence of lifecycle refinement hooks instead of directly returning the raw multimodal report.
As shown in Figure~\ref{fig:main}, this process consists of six steps: (1) \textit{Section Refine} revises each section for clarity, evidence coverage, citation fidelity, and local coherence; (2) \textit{Image Refine} decides whether each visual element should be \texttt{Keep}, \texttt{Delete}, or \texttt{Edit}, and executes editing instructions for images marked as \texttt{Edit}; (3) \textit{Overall Refine} improves global organization, cross-section consistency, and image--text alignment; (4) \textit{HTML Generate} converts the refined report into an HTML document with layout and styling specifications; (5) \textit{HTML Refine} further adjusts the HTML structure, style consistency, spacing, and rendered readability; and (6) \textit{Render} displays the final HTML document in a browser as a user-facing multimodal report.
Together, these refinement and rendering steps improve the readability and usability of the final report by presenting its layout, visual placement, and image--text organization in a form that is directly accessible to human readers.

%% End Ptah Method

\begin{figure}[!t]
    \centering
    \includegraphics[width=\linewidth]{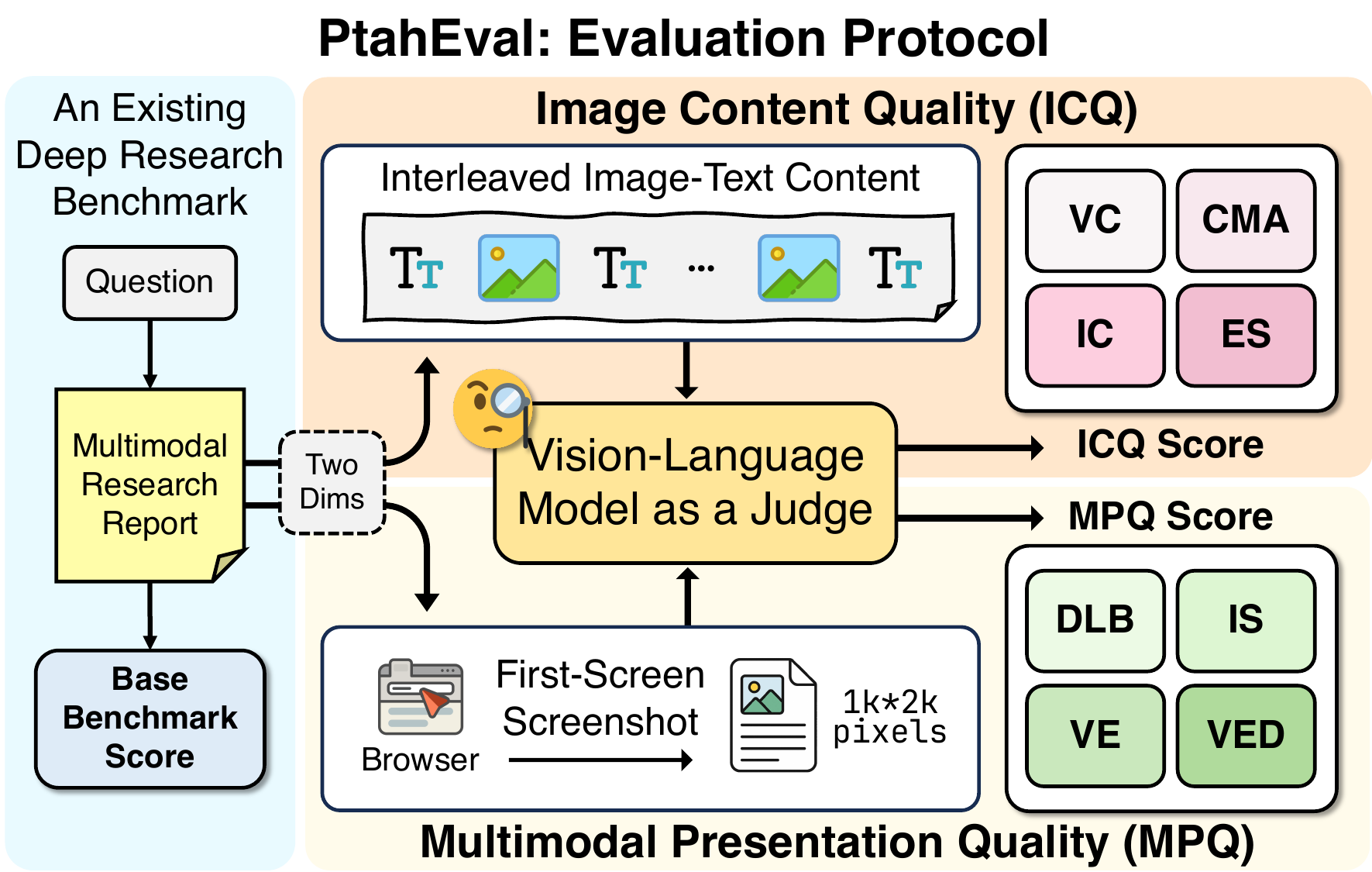}
    \vspace{-0.5em}
    \caption{An illustration of our \textsc{Ptah}Eval evaluation.}
    \vspace{-0.5em}
    \label{fig:ptaheval}
\end{figure}

\begin{table*}[t]
    \centering
    \caption{The overall results on DeepResearch Bench and DeepConsult. DeepResearch Bench evaluates Comprehensiveness (Comp.), Insight/Depth (Insight), Instruction-Following (Inst.), and Readability (Read.), together with the overall score. DeepConsult evaluates Instruction-Following (Inst.), Comprehensiveness (Comp.), Completeness (Compl.), and Writing Quality (Writ.), along with their average score.}
    \small
    \begin{tabularx}{\textwidth}{l *{10}{>{\centering\arraybackslash}X}}
    \toprule
    \multirow{2}{*}{\textbf{Method}}
     & \multicolumn{5}{c}{\textbf{DeepResearch Bench}}
     & \multicolumn{5}{c}{\textbf{DeepConsult}} \\
    \cmidrule(lr){2-6}
    \cmidrule(lr){7-11}
     & Comp. & Insight & Inst. & Read. & Overall & Inst. & Comp. & Compl. & Writ. & AVG. \\
    \midrule
    
    \multicolumn{11}{l}{\textbf{\textit{Direct Generation}}} \\

    Qwen3-32B
    & 40.73 & 39.59 & 45.85 & 44.80 & 42.22 
    & 0.98 & 0.98 & 0.98 & 0.98 & 0.98 \\

    QwQ-32B 
    & 40.97 & 40.27 & 46.09 & 45.23 & 42.59 
    & 0.98 & 0.98 & 0.98 & 1.96 & 1.23 \\
    
    \midrule
    
    \multicolumn{11}{l}{\textbf{\textit{Text-Only Generation}}} \\

    ReAct 
    & 42.63 & 40.42 & \textbf{47.66} & 46.37 & 43.70 
    & 0.98 & 0.98 & 0.98 & 6.86 & 2.45 \\

    Search-o1 
    & 41.57 & 39.70 & 46.82 & 45.96 & 42.91 
    & 0.98 & 0.98 & 0.98 & \underline{7.84} & 2.69 \\

    WebThinker 
    & \textbf{44.63} & \underline{43.26} & \underline{46.86} & \underline{46.61} & \underline{45.00} 
    & \underline{2.94} & \underline{17.64} & \underline{2.94} & 5.88 & \underline{7.35} \\

    \midrule
    
    \multicolumn{11}{l}{\textbf{\textit{Multimodal Generation}}} \\

    LLM-I 
    & 35.14 & 31.77 & 41.14 & 40.07 & 36.36 
    & 0.98 & 0.98 & 0.98 & 1.96 & 1.23 \\

    \midrule
\rowcolor[RGB]{236,244,252}
    \textbf{\textsc{Ptah} (ours)} 
    & \underline{42.97} & \textbf{44.32} & 46.71 & \textbf{47.95} & \textbf{45.16} 
    & \textbf{13.73} & \textbf{18.63} & \textbf{17.64} & \textbf{14.71} & \textbf{16.18} \\

    \bottomrule
    \end{tabularx}
    \label{tab:main_results}
\end{table*}

\section{\textsc{Ptah}Eval Evaluation Protocol}

Existing evaluation protocols for deep research systems focus mainly on textual outputs and are insufficient for multimodal reports that integrate textual arguments, visual evidence, and rendered layouts. We propose \textsc{Ptah}Eval, a flexible protocol that preserves the original questions and text-oriented metrics of existing benchmarks while adding multimodal evaluation procedures over the generated report artifact. Given a benchmark query, a system must produce a rendered multimodal report rather than a text-only answer, which is then assessed from two complementary perspectives: \textit{Image Content Quality} (ICQ), measuring whether individual images are clear, relevant, informative, and aligned with the surrounding text; and \textit{Multimodal Presentation Quality} (MPQ), measuring whether the rendered report presents information in a readable, well-organized, and visually coherent manner.

\subsection{Image Content Quality Evaluation}

For ICQ, we feed interleaved text--image inputs to a VLM, which judges whether each image meaningfully contributes to the report in terms of informativeness, consistency with the surrounding text, and support for textual explanations. ICQ comprises four dimensions:
(1) \textbf{Visual Clarity (VC)}: image legibility and ease of interpretation;
(2) \textbf{Cross-Modal Alignment (CMA)}: semantic consistency with the surrounding text and appropriateness of the placement context;
(3) \textbf{Information Complementarity (IC)}: whether the image conveys meaningful information that complements the text, especially content hard to express in words alone;
(4) \textbf{Evidentiary Support (ES)}: whether the image supports, explains, or contextualizes the claims and conclusions in the surrounding text.

\subsection{Multimodal Presentation Quality Evaluation}

MPQ targets the presentation quality of the rendered report under realistic reading conditions. Since \textsc{Ptah} produces a user-facing web artifact, the report is first rendered as a webpage and its visible viewport ($1000 \times 2000$ pixels) is captured as the evaluation input, reflecting what human readers see in terms of layout, spacing, visual placement, and image--text organization. The captured page image is then assessed by the VLM along four dimensions:
(1) \textbf{Density-Legibility Balance (DLB)}: balance between information density and perceptual clarity within the viewport;
(2) \textbf{Informational Saliency (IS)}: whether key insights and structural elements are effectively highlighted through visual hierarchy;
(3) \textbf{Visual Encoding Diversity (VED)}: use of diverse visual forms (e.g., tables, callouts, icons, charts, diagrams, illustrative figures) to support comprehension;
(4) \textbf{Visual Ergonomics (VE)}: spacing, visual rhythm, alignment, and perceptual clarity, evaluating whether the layout reduces reading effort while preserving clear entry points.

Following \citet{seongyun2024prometheus}, each ICQ and MPQ dimension is scored on a five-point Likert scale (1--5). Together with the original benchmark metrics, ICQ and MPQ provide complementary signals on textual reliability, image-level quality, and report-level presentation.

\section{Experiments}

\subsection{Experimental Setup}

\paragraph{Implementation.}
We use Qwen3-32B~\cite{tongyi2025qwen3} as the Planner, Researcher, and Verifier, and Qwen3-VL-32B-Instruct~\cite{bai2025qwen3vl} as the Writer. 
Qwen3-32B is additionally employed for LLM-based verification, while Qwen3-VL-32B-Instruct is used for image selection during the Research stage. 
Detailed descriptions of all tools are provided in Appendix~\ref{sec:tools}.

\paragraph{Datasets and Baselines.}
We use the widely adopted benchmark \textbf{DeepResearch Bench}~\cite{du2025drb}. 
Following \citet{han2025tdd-dr}, we additionally include \textbf{DeepConsult}~\cite{youdotcom2025deepconsult}. 
We generate reports using questions from both benchmarks and evaluate the textual content using the evaluation metrics defined in each benchmark. 
To accommodate interleaved text–image outputs, we replace all LLM-as-judge evaluators with Qwen3-VL-235B-A22B-Instruct, a VLM capable of jointly processing textual and visual inputs.

As baselines, we include two direct report generation methods using Qwen3-32B and QwQ-32B. 
We also compare with three single-agent text-only search methods: ReAct~\cite{yao2023react}, Search-o1~\cite{li2025searcho1}, and WebThinker~\cite{li2025webthinker}. 
Since there is currently no readily reproducible open-source general multimodal research report generation framework, we additionally include LLM-I~\cite{guo2025llm-i}, an agent-based method for generating multimodal content, as a baseline. All these baselines use Qwen3-32B as the base model. For all methods, we uniformly retrieve the top-5 web pages for each query as external knowledge during text search.

\begin{table*}[t]
    \centering
    \small
    \caption{Overall \textsc{Ptah}Eval results on DeepResearch Bench. The best results are highlighted in \textbf{bold}, and the second-best results are \underline{underlined}. Since direct generation and text-only generation baselines do not produce reports with images, their \textit{Image Content Quality} scores are not applicable and are marked as ``-''.}
    \begin{tabularx}{\textwidth}{
        l
        *{4}{>{\centering\arraybackslash}X}
        >{\columncolor{gray!10}\centering\arraybackslash}X
        *{4}{>{\centering\arraybackslash}X}
        >{\columncolor{gray!10}\centering\arraybackslash}X
    }
    \toprule
    \multirow{2}{*}{\textbf{Method}}
     & \multicolumn{5}{c}{\textbf{Image Content Quality}}
     & \multicolumn{5}{c}{\textbf{Multimodal Presentation Quality}} \\
    \cmidrule(lr){2-6}
    \cmidrule(lr){7-11}
     & VC & CMA & IC & ES & \textbf{Avg.} & DLB & IS & VED & VE & \textbf{Avg.} \\
    \midrule
    
    \multicolumn{11}{l}{\textbf{\textit{Direct Generation}}} \\
    Qwen3-32B & - & - & - & - & - & 3.55 & 3.60 & 2.73 & 3.53 & 3.35 \\
    QwQ-32B   & - & - & - & - & - & \underline{3.65} & \underline{3.67} & 2.84 & \underline{3.65} & \underline{3.45} \\
    
    \midrule
    
    \multicolumn{11}{l}{\textbf{\textit{Text-Only Generation}}} \\
    ReAct      & - & - & - & - & - & 3.13 & 3.21 & 2.41 & 3.55 & 3.08 \\
    Search-o1  & - & - & - & - & - & 3.49 & 3.14 & 2.62 & 3.24 & 3.12 \\
    WebThinker & - & - & - & - & - & 3.42 & 2.89 & 2.78 & 3.35 & 3.11 \\

    \midrule
    
    \multicolumn{11}{l}{\textbf{\textit{Multimodal Generation}}} \\
    LLM-I & \underline{2.10} & \underline{2.28} & \underline{1.96} & \underline{1.52} & \underline{1.97} & 3.12 & 2.51 & \underline{3.25} & 3.11 & 3.00 \\

    \midrule
    \rowcolor[RGB]{236,244,252}
    \textbf{\textsc{Ptah} (ours)} & \textbf{4.42} & \textbf{4.79} & \textbf{4.35} & \textbf{4.01} & \textbf{4.39} & \textbf{3.72} & \textbf{3.78} & \textbf{3.61} & \textbf{3.74} & \textbf{3.71} \\

    \bottomrule
    \end{tabularx}
    \label{tab:ptaheval_results}
\end{table*}

\subsection{Main Results}

We evaluate \textsc{Ptah} against baselines along three dimensions: textual content quality, visual quality, and factual credibility. \textsc{Ptah} consistently surpasses prior approaches in all three.

\textbf{(1) Overall Content Quality.} Table~\ref{tab:main_results} reports results on DeepResearch Bench and DeepConsult. On DeepResearch Bench, \textsc{Ptah} attains the best overall score of 45.16, leading on \textit{Insight/Depth} and \textit{Readability} while remaining competitive on \textit{Instruction-Following} and \textit{Comprehensiveness}, demonstrating that the multi-agent decomposition yields reports with deeper analysis and clearer structural organization. On DeepConsult, \textsc{Ptah} outperforms every baseline across all dimensions, reaching an average of \textbf{16.18}---more than double the best baseline (WebThinker). The most pronounced gains, on \textit{Instruction-Following}, \textit{Completeness}, and \textit{Writing Quality}, indicate stronger adherence to complex task specifications.

\textbf{(2) Visual Quality.} Table~\ref{tab:ptaheval_results} reports \textsc{Ptah}Eval results on DeepResearch Bench. For \textit{Image Content Quality}, \textsc{Ptah} achieves the highest scores across all four dimensions. Its near-ceiling \textit{Cross-Modal Alignment} score stems from two factors: (i) retrieved webpage images are inherently aligned with their surrounding textual context, and (ii) the test-time scaling (TTS) mechanism further refines image--text coherence. By contrast, the multimodal baseline LLM-I performs markedly worse, confirming that images produced by \textsc{Ptah} are clearer, more contextually relevant, and more effective as supporting evidence.

For \textit{Multimodal Presentation Quality}, \textsc{Ptah} likewise leads on every dimension, benefiting from improved image quality and TTS-driven HTML layout optimization. Gains on \textit{Density-Legibility Balance} and \textit{Visual Ergonomics} reflect well-balanced spacing, stronger visual rhythm, and reduced perceptual load, while gains on \textit{Informational Saliency} and \textit{Visual Encoding Diversity} show that \textsc{Ptah} anchors images in appropriate contexts and leverages diverse visual forms to highlight key insights and guide reader attention. Unlike text-only baselines, which rely on plain prose, or LLM-I, which lacks systematic layout refinement, \textsc{Ptah} integrates visual elements into a coherent global page design, yielding more professional and readable multimodal reports.

\textbf{(3) Credibility.} Table~\ref{tab:fact_results} reports FACT metrics on DeepResearch Bench. Since open-source baselines do not natively generate references, we prompt them to produce references alongside their reports. \textsc{Ptah} attains a Citation Accuracy of 87.53 with 9.64 effective citations per task, substantially outperforming all baselines. Case studies show that baselines frequently emit invalid or hallucinated URLs, whereas the Verifier Agent guarantees that every reference in \textsc{Ptah} maps to a valid, accessible source; the residual errors mainly stem from minor mismatches between cited content and the corresponding source. \textsc{Ptah} also issues more search tool calls than competing methods, reflecting its more thorough exploration of external knowledge, which directly translates into stronger factual grounding.

\begin{figure*}
    \centering
    \includegraphics[width=\textwidth]{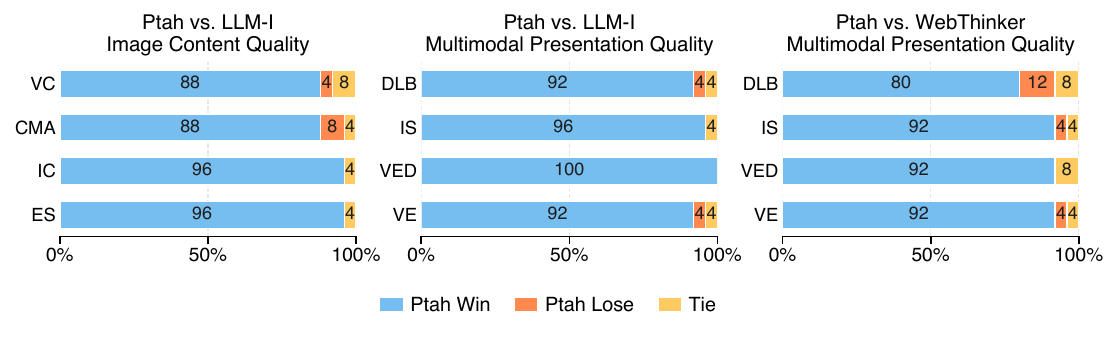}
    \vspace{-2em}
    \caption{Human evaluation of \textsc{Ptah} against LLM-I and WebThinker on DeepResearch Bench via \textsc{Ptah}Eval.}
    \label{fig:human}
\end{figure*}

\subsection{Human Evaluation}

To validate the VLM-based judgments used in \textsc{Ptah}Eval, we randomly sample 25 tasks from DeepResearch Bench and collect reports from \textsc{Ptah}, LLM-I, and WebThinker. Four annotators (two Ph.D. and two undergraduate students in AI) perform anonymized pairwise comparisons between \textsc{Ptah} and each baseline under the \textsc{Ptah}Eval criteria, with majority voting determining the final preference.
As shown in Figure~\ref{fig:human}, human preferences closely track the VLM-based scores. Annotators consistently favor \textsc{Ptah} over LLM-I on Image Content Quality, indicating clearer images that better complement and support the surrounding text, and over both LLM-I and WebThinker on Multimodal Presentation Quality, indicating that the gains stem from coherent multimodal organization rather than superficial image insertion. These findings confirm both the reliability of the \textsc{Ptah}Eval evaluator and the consistent advantage of \textsc{Ptah}.

\begin{table}[t]
\centering
\small
\caption{FACT evaluation on DeepResearch Bench. We report Citation Accuracy (C.Acc.) and the average number of Effective Citations per task (E.Cit.), along with the average number of search tool calls per task.}
\begin{tabular}{lccc}
\toprule
\textbf{Method} & \textbf{C.Acc.} &\textbf{E.Cit.} & \textbf{\#Search} \\
\midrule

ReAct       & 37.28 & 0.23 & 4.17 \\
Search-o1   & 40.91 & 0.31 & 2.78 \\
WebThinker  & 60.74 & 2.32 & 5.91 \\
\midrule
\rowcolor[RGB]{236,244,252}
\textsc{Ptah} w/o Verifier & 30.29 & 4.75 & 5.13 \\
\rowcolor[RGB]{236,244,252}
\textsc{Ptah} & 87.53 & 9.64 & 12.82 \\

\bottomrule
\end{tabular}
\label{tab:fact_results}
\end{table}

\subsection{Ablation Studies}

\paragraph{Influence of the Verifier Agent.}
We remove the Verifier and rerun generation on DeepResearch Bench. Without it, 14 out of 100 tasks fail to proceed in the planning stage due to parsing errors, repetitive outputs, or incorrect tool calls. Among the remaining 86 tasks, another 18 fail during the research stage, leaving only 68 tasks that successfully produce final reports. This highlights the role of the Verifier in maintaining the stability of the multi-agent framework.

We further evaluate these 68 reports under the FACT metrics and record the number of text-search tool calls (Table~\ref{tab:fact_results}). Removing the Verifier causes substantial drops in both citation validity and factual correctness. The comparison also shows that Verifier feedback encourages the model to issue additional search calls, thereby expanding its exploration of external knowledge.

\paragraph{Effect of Test-Time Scaling.}
We further study the impact of test-time scaling in Table~\ref{tab:tts_ablation}. Removing TTS reduces the overall DRB score by 3.03 points, and Image Content Quality and Multimodal Presentation Quality also decline noticeably. This indicates that TTS plays an important role in improving content quality, image quality, and the final HTML rendering of the report.

We additionally observe that without TTS the model inserts more invalid images and exhibits a higher rate of image-generation failures, further confirming that TTS is crucial for producing stable, high-quality multimodal reports.

\begin{table}[t]
\centering
\small
\caption{Ablation on test-time scaling (TTS), reporting the overall DeepResearch Bench (DRB) score, average Image Content Quality (ICQ) and Multimodal Presentation Quality (MPQ) scores, and the average number of generated and failed images per report.}
\begin{tabular}{lcccc}
\toprule
\textbf{Method} & \textbf{DRB} &\textbf{ICQ} & \textbf{MPQ} & \textbf{\#Img / \#Fail} \\
\midrule

LLM-I                 & 36.36 & 1.97 & 3.00 & 0.74 / 0.14 \\
\midrule
\rowcolor[RGB]{236,244,252}
\textsc{Ptah} w/o TTS & 42.13 & 2.77 & 3.49 & 5.06 / 0.38 \\
\rowcolor[RGB]{236,244,252}
\textsc{Ptah}         & 45.16 & 4.39 & 3.71 & 3.76 / 0.12 \\

\bottomrule
\end{tabular}
\label{tab:tts_ablation}
\end{table}

\section{Conclusion}

We present \textsc{Ptah}, a multi-agent harness for verifiable multimodal deep research that addresses the lack of stage-wise verification and the post-hoc use of visual evidence in existing systems. 
\textsc{Ptah} decomposes the research lifecycle from query to rendered report into \textit{Planning}, \textit{Research}, and \textit{Writing}, where a \textit{Verifier Agent} enforces factual grounding, citation fidelity, and cross-modal consistency through rule-based and LLM-based checks across all stages. 
Together with \textsc{Ptah}Eval, which augments existing benchmarks with image content and presentation metrics, our experiments, human studies, and ablations show that \textsc{Ptah} consistently produces credible and professionally interleaved reports. 
These results advance multimodal deep research toward evidence-grounded, visually informative, and human-centric report generation.

\section*{Limitations}

Due to the constrained reasoning capabilities of existing open-source models, achieving a stable and autonomous agentic workflow for long-horizon multimodal search and generation remains a significant challenge. To ensure system controllability and reliability, we decompose the holistic framework into three distinct sequential stages rather than adopting a single-pass agentic generation process. While this modular design introduces manually defined boundaries, it facilitates more granular monitoring and rigorous validation of intermediate outputs. Furthermore, this decoupled architecture allows for independent optimization of specific modules in future iterations.

\bibliography{ref}

\appendix

\section{More Implementation Details}

All experiments were conducted on a machine equipped with 4 $\times$ A800 80GB GPUs. On this machine, we locally deployed Qwen3-32B and Qwen3-VL-32B-Instruct using vLLM. For image generation, image editing, and evaluation, we accessed Qwen-Image, Qwen-Image-Edit, and Qwen3-VL-235B-A22B-Instruct through the APIs provided by the SiliconFlow platform\footnote{SiliconFlow: \url{https://www.siliconflow.cn}}. Text search and image search APIs were provided by Serper\footnote{Serper: \url{https://serper.dev}}, and webpage parsing was supported by Jina Reader\footnote{Jina Reader: \url{https://jina.ai/reader}}.

External APIs were used as replaceable implementation interfaces in our experiments. For image generation, image editing, and VLM-based evaluation, the API-accessed components correspond to open-source models. These models were invoked with their default inference parameters unless otherwise specified. The use of hosted APIs was mainly intended to reduce the local GPU cost of large-scale experiments; the same components can be reproduced by locally deploying the corresponding open-source models with the same settings.

For web access, Serper and Jina Reader serve as general-purpose interfaces for search result retrieval and webpage parsing, respectively. This follows the common setting of web-augmented deep research and search-based agent systems, where agents interact with external search and reading tools to obtain up-to-date evidence. These interfaces are not tied to the core design of \textsc{Ptah} and can be replaced by alternative search engines, browser tools, or local retrieval systems.

\section{Details of the Tools}
\label{sec:tools}
To support the distinct operational needs of our multi-agent framework, we integrate a suite of specialized tools. The \textsc{Planner} and \textsc{Researcher} leverage text retrieval to gather information, while the \textsc{Writer} utilizes visual retrieval, synthesis, and code execution for draft composition, supported by image editing for subsequent refinement. The specific implementations are defined as follows:

\begin{itemize}[leftmargin=1em]
    \item \textbf{Text Search:} Facilitates knowledge acquisition. Accepting a text query as input, this tool returns summaries of the top-$K$ relevant web pages. The pipeline first retrieves URLs via the Google Search API, parses the raw HTML into Markdown using the Jina Reader API\footnote{Jina Reader: \url{https://jina.ai/reader}}, and finally employs a Qwen3-32B model to summarize each page for key information extraction.
    
    \item \textbf{Image Search:} Retrieves factual visual evidence, such as specific real-world entities. It maps a text query to the top-$K$ matching images using the Google Image Search API (via Serper).
    
    \item \textbf{Image Generation:} Synthesizes thematic illustrations or abstract concepts. It converts textual descriptions into images using the Qwen-Image diffusion model~\cite{tongyi2025qwenimage}, hosted on the SiliconFlow API.
    
    \item \textbf{Image Editing:} Modifies visual details or adjusts styles during the refinement phase. It accepts an initial image (whether retrieved or generated) alongside a textual editing instruction to produce a modified output via the Qwen-Image-Edit-2509 model~\cite{tongyi2025qwenimage}.
    
    \item \textbf{Code Execution:} Enables precise data visualization. It executes generated Python scripts within a secure, isolated sandbox environment to render rigorous charts and plots from structured data.
\end{itemize}

\section{Dataset Details}
\paragraph{DeepResearch Bench.}
DeepResearch Bench is a comprehensive benchmark for evaluating deep research agents on complex, long-form analytical tasks.  It consists of 100 PhD-level research tasks spanning 22 distinct domains, with 50 tasks in English and 50 in Chinese, each carefully designed and curated by domain experts and senior practitioners to ensure high standards of complexity, clarity, and realism.  Each task requires generating a comprehensive research report that involves multi-step web exploration, information integration, and analytical reasoning. The benchmark provides a diverse and realistic evaluation setting for assessing models' capabilities in deep research, covering both report generation quality and information retrieval effectiveness.

\paragraph{DeepConsult.}
DeepConsult is a benchmark designed for evaluating deep research capabilities on business and consulting-oriented queries. It contains 102 queries covering a broad range of real-world consulting scenarios, including market analysis, investment opportunity assessment, industry evaluation, financial modeling, technology trend analysis, and strategic business planning. Each query is formulated to require comprehensive analysis and multi-step reasoning, reflecting the complexity of practical consulting tasks. The dataset is designed to assess whether models can produce structured, insightful, and actionable reports comparable to those generated in professional consulting settings. 

\begin{figure*}[!t]
    \centering
    \includegraphics[width=\linewidth]{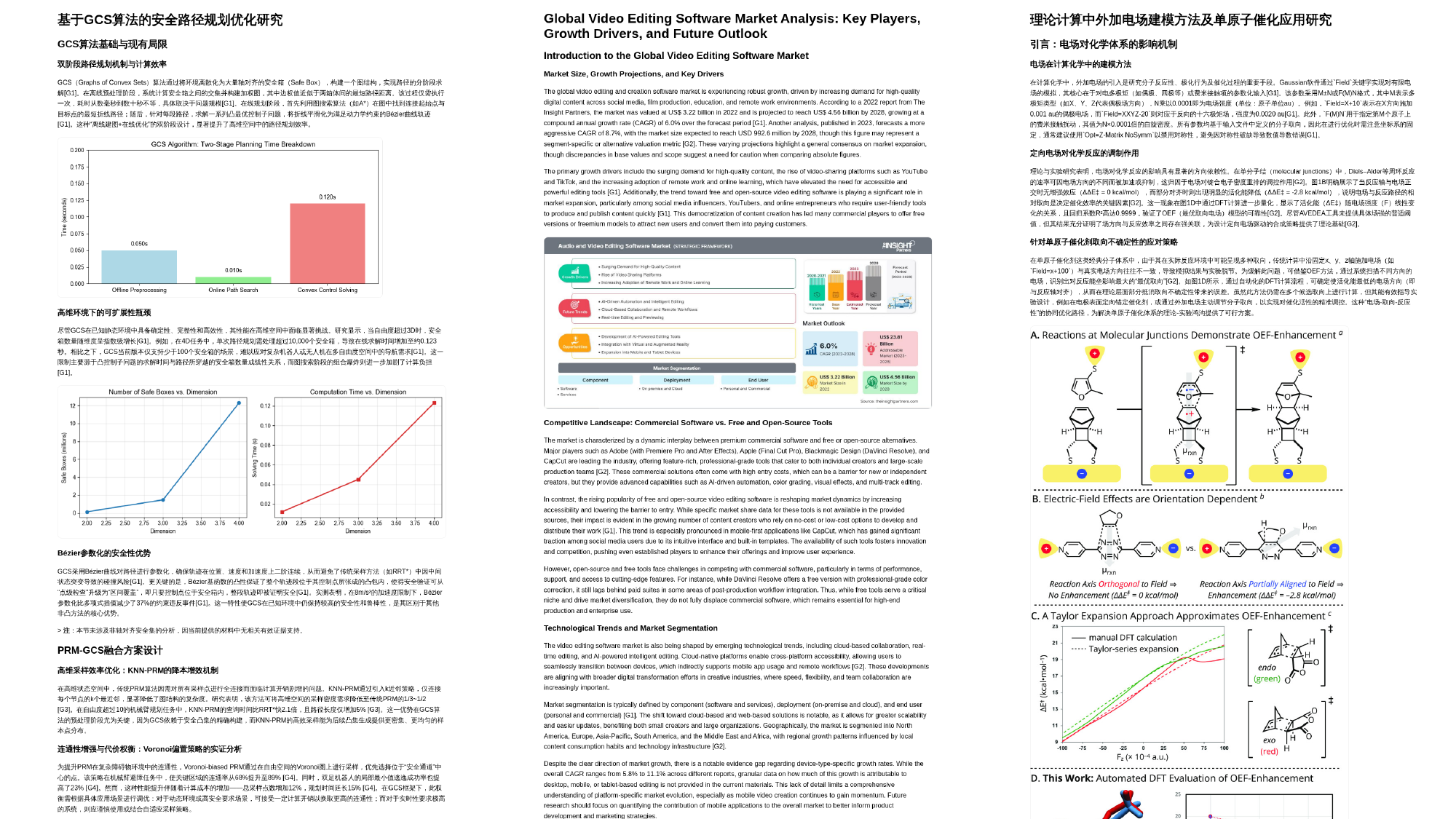}
    \caption{First-screen views of multimodal analytical reports generated by the \textsc{Ptah} framework.}
    \label{fig:example}
\end{figure*}

\section{Verifier Agent Details}
\label{sec:verifier}

The Verifier agent operates across the Planning, Research, and Writing stages, providing both rule-based and LLM-based verification to ensure the correctness, consistency, and quality of the generated outputs.

\paragraph{Planning Stage.}
During the planning stage, rule-based verification primarily checks the structural validity of the Planner's outputs, including output formatting and the correctness of tool invocation schemas. 
In parallel, LLM-based rubric verification evaluates the entire reasoning trajectory of the Planner, including intermediate thoughts, tool calls, tool responses, and the final plan. 
It assesses aspects such as the rationality of the search strategy, the completeness of the generated outline, and the consistency between the outline and the retrieved web content. 
The Verifier produces a structured evaluation consisting of a scoring rubric and a review report, which summarizes strengths, weaknesses, potential improvements, and a final decision (accept or reject).

\paragraph{Research Stage.}
In the research stage, rule-based verification plays a critical role in ensuring citation fidelity. Specifically, all referenced URLs in the final output must exactly match those retrieved and accessed during the text search process, guaranteeing strict consistency between citations and evidence sources. 
It also validates the structural correctness of the output format. 
Meanwhile, LLM-based rubric verification focuses on assessing the depth and completeness of the exploration process, as well as the consistency between the synthesized findings and the retrieved web content, ensuring both the rigor and reliability of the research.

\paragraph{Writing Stage.}
During the writing stage, the Verifier guides the Writer in refining and polishing the generated report. 
It ensures the correctness of image tool invocation syntax and enforces consistency between the final report and the research findings. 
Through iterative feedback, the Verifier helps improve both the clarity and the fidelity of the generated content.

\section{Details of Webpage Image Selection}

During the research stage, the Researcher actively invokes the text search tool. 
Through this process, Jina Reader extracts the full content of retrieved web pages, including the URLs of embedded images. 
We first download all accessible images from these web pages.

To ensure image quality, we apply a rule-based filtering step to remove low-quality or irrelevant images. 
Specifically, we discard images with low resolution, extremely small dimensions, extreme aspect ratios, as well as SVG files. 
This step effectively filters out non-informative visuals such as logos, icons, and banner images.

The remaining images are then processed in batches by a VLM. 
Conditioned on the Planner's specified image requirements, as well as its own assessment of image quality and semantic relevance, the VLM determines whether each image should be retained or discarded.

Finally, for each section, we construct a curated \textit{Visual Working Memory}, which serves as the candidate image set for the subsequent writing stage.

\section{Efficiency Analysis}

We further analyze the computational efficiency of \textsc{Ptah} on DeepResearch Bench. 
Table~\ref{tab:stage-wise-latency} reports the average wall-clock latency of each stage. 
The full pipeline takes 1015 seconds on average. 
Among all stages, Research is the most time-consuming stage, taking 459 seconds on average, because it requires open-ended evidence collection, webpage inspection, and image-pool construction for multiple report sections. 
Test-Time Scaling (TTS) takes 243 seconds on average, reflecting the additional cost of verifier-guided section refinement, image refinement, overall refinement, and HTML refinement. 
These results show that \textsc{Ptah} introduces additional computation compared with text-only deep-research agents, but the cost is mainly concentrated in evidence acquisition and final multimodal refinement, which are essential for credible multimodal report generation.

\begin{table}[t]
\centering
\small
\begin{tabular}{lc}
\toprule
\textbf{Stage} & \textbf{Avg. Time (s)} \\
\midrule
Planning Stage & 192 \\
Research Stage & 459 \\
Writing Stage & 121 \\
TTS & 243 \\
\midrule
Total & 1015 \\
\bottomrule
\end{tabular}
\caption{
Stage-wise average latency of \textsc{Ptah} on DeepResearch Bench.
Research-stage latency is measured as the wall-clock time of parallel section-level investigations.
}
\label{tab:stage-wise-latency}
\end{table}

We also evaluate the efficiency benefit of parallel section-level research. 
As shown in Table~\ref{tab:parallel-research-latency}, executing Researchers in parallel reduces the average Research-stage latency from 1328 seconds to 459 seconds. 
This corresponds to a 65.4\% reduction in wall-clock time, or a 2.89$\times$ slowdown when the same Researcher agents are executed sequentially. 
The result indicates that the multi-agent design of \textsc{Ptah} is not merely an added source of computation; it also serves as an important efficiency mechanism by decomposing a long-form research task into section-level investigations that can be performed concurrently.

\begin{table}[t]
\centering
\small
\begin{tabular}{lcc}
\toprule
\textbf{Research Execution} & \textbf{Avg. Time (s)} &\textbf{ Relative Change} \\
\midrule
Parallel & 459 & 1.00$\times$ \\
Sequential & 1328 & 2.89$\times$ slower \\
\bottomrule
\end{tabular}
\caption{
Latency comparison between parallel and sequential Researcher execution.
Parallel section-level research substantially reduces the wall-clock latency of the Research stage.
}
\label{tab:parallel-research-latency}
\end{table}

Finally, we examine the latency impact of verifier strength. 
Table~\ref{tab:verifier-latency} compares the current Verifier with a stronger reasoning model, DeepSeek-R1. 
Replacing the current Verifier with DeepSeek-R1 increases Planning latency from 192 seconds to 853 seconds and Research latency from 459 seconds to 1408 seconds. 
This increase comes from both the longer reasoning time of the stronger verifier and the additional revision rounds triggered by stricter verification. 
Therefore, verifier selection introduces a clear quality--efficiency trade-off: stronger verifiers may provide stricter intermediate checking, but they can substantially increase the overall latency of long-form multimodal report generation. 
In our main experiments, we use the current Verifier setting as a balanced configuration that preserves strong factual grounding and cross-modal consistency while avoiding excessive revision overhead.

\begin{table}[t]
\centering
\small
\begin{tabular}{lc}
\toprule
\textbf{Setting} & \textbf{Time (s)} \\
\midrule
Current Verifier -- Planning & 192 \\
Current Verifier -- Research & 459 \\
DeepSeek-R1 Verifier -- Planning & 853 \\
DeepSeek-R1 Verifier -- Research & 1408 \\
\bottomrule
\end{tabular}
\caption{
Latency impact of verifier strength.
A stronger reasoning verifier substantially increases both Planning and Research latency due to more expensive verification and additional revision rounds.
}
\label{tab:verifier-latency}
\end{table}

\subsection{User-Centric Human Evaluation}

We conduct a user-centric human evaluation to further examine the practical reading experience of multimodal deep-research reports. 
While \textsc{Ptah}Eval provides a scalable automatic protocol for evaluating image content quality and multimodal presentation quality, human evaluation offers complementary evidence on whether the generated reports are readable, usable, and helpful for information acquisition in realistic viewing scenarios.

We randomly sample 20 reports from DeepResearch Bench and compare the rendered HTML reports generated by \textsc{Ptah} and WebThinker. 
WebThinker is selected as the comparison system because it is the strongest text-only deep-research baseline in our experiments. 
We recruit four evaluators, including two AI PhD students as expert evaluators and two AI undergraduate students as general users. 
Each evaluator is asked to compare the two reports for the same query and judge whether \textsc{Ptah} wins, ties, or loses against WebThinker along four user-centric dimensions: readability, usability, information acquisition efficiency, and overall preference. 
Readability measures whether the report is easy to read and visually clear. 
Usability measures whether the report organization supports convenient browsing and understanding. 
Information acquisition efficiency measures whether users can quickly identify and absorb key information from the report. 
Overall preference captures the evaluator's holistic judgment of which report better supports the research task.

Table~\ref{tab:user-centric-human-eval} reports the win-or-tie rate of \textsc{Ptah} over WebThinker. 
\textsc{Ptah} achieves high win-or-tie rates across all four dimensions, with 88.75\% on readability, 88.75\% on usability, 96.25\% on information acquisition efficiency, and 95.00\% on overall preference. 
These results indicate that the multimodal reports generated by \textsc{Ptah} are not only preferred by automatic evaluators, but also provide a better practical reading experience for human users. 
In particular, the large gain in information acquisition efficiency suggests that interleaved visual evidence helps users locate and understand important information more effectively than text-only reports. 
This human evaluation complements \textsc{Ptah}Eval by showing that the improvements in multimodal presentation quality correspond to meaningful gains in perceived usability and reading efficiency.

\begin{table}[t]
\centering
\small
\begin{tabular}{lcccc}
\toprule
\textbf{Evaluator} & \textbf{Read.} & \textbf{Usability} & \textbf{Info.} & \textbf{Overall} \\
\midrule
Expert E1 & 85\% & 90\% & 95\% & 95\% \\
Expert E2 & 85\% & 80\% & 95\% & 90\% \\
General U1 & 90\% & 95\% & 100\% & 100\% \\
General U2 & 95\% & 90\% & 95\% & 95\% \\
\midrule
Average & 88.75\% & 88.75\% & 96.25\% & 95.00\% \\
\bottomrule
\end{tabular}
\caption{
User-centric human evaluation of \textsc{Ptah} against WebThinker on 20 sampled DeepResearch Bench reports.
Each value denotes the win-or-tie rate of \textsc{Ptah}.
``Info.'' denotes information acquisition efficiency and ``Read.'' denotes readability.
}
\label{tab:user-centric-human-eval}
\end{table}

\subsection{Ablation on Visual Elements}

We conduct an additional same-framework ablation to isolate the contribution of visual elements in \textsc{Ptah}. 
The ablated variant, denoted as \textsc{Ptah} w/o images, uses the same Planning--Research--Writing pipeline, stage-wise verification, and test-time refinement procedure as the full \textsc{Ptah} system, but removes all images from the final rendered report. 
This setting allows us to examine whether the improvement comes from the multimodal visual elements themselves or only from the underlying agentic research framework.

Table~\ref{tab:visual-elements-ablation} reports the results on DeepResearch Bench. 
Removing images only slightly changes the text-oriented DRB overall score, from 45.16 to 45.10. 
This is expected because DeepResearch Bench primarily evaluates textual research quality and does not directly reward whether visual evidence is relevant, evidential, or helpful for reading. 
Importantly, \textsc{Ptah} w/o images still outperforms WebThinker on DRB overall, indicating that the Planning--Research--Writing framework and stage-wise verification preserve strong textual research quality even without visual output.

In contrast, removing images leads to a clear drop in multimodal presentation quality. 
The MPQ average score decreases from 3.71 to 3.29 after visual elements are removed. 
This result shows that interleaved visuals make a substantial contribution to the rendered report's multimodal presentation quality, rather than merely changing its surface appearance. 
Compared with WebThinker, the full \textsc{Ptah} system improves both DRB overall and MPQ average, suggesting that \textsc{Ptah} enhances multimodal readability and presentation while maintaining competitive textual research quality.

\begin{table}[t]
\centering
\small
\begin{tabular}{lcc}
\toprule
\textbf{Method} & \textbf{DRB Overall} & \textbf{MPQ Avg.} \\
\midrule
WebThinker & 45.00 & 3.11 \\
\textsc{Ptah} w/ images & 45.16 & 3.71 \\
\textsc{Ptah} w/o images & 45.10 & 3.29 \\
\bottomrule
\end{tabular}
\caption{
Ablation on visual elements.
\textsc{Ptah} w/o images keeps the same framework as \textsc{Ptah}, but removes all images from the final report.
}
\label{tab:visual-elements-ablation}
\end{table}

\section{Example Cases}

As shown in Figure~\ref{fig:example}, we present three representative first-screen views of multimodal analytical reports generated by the \textsc{Ptah} framework.

\section{Usage of AI Assistants}
The authors used ChatGPT only for language polishing and grammar correction during the preparation of this manuscript. All research content and technical contributions were developed and verified by the authors.

\end{document}